\pdfoutput=1

\documentclass[11pt]{article}

\usepackage{acl}

\usepackage{times}
\usepackage{latexsym}

\usepackage[T1]{fontenc}

\usepackage[utf8]{inputenc}

\usepackage{microtype}

\usepackage{inconsolata}
\usepackage{multirow}

\usepackage{graphicx}
\usepackage{tikz}
\usepackage{subcaption}
\usepackage{float}
\usepackage[normalem]{ulem}
\usepackage{soul}
\usepackage{multicol}
\usepackage{pgfplots}
\pgfplotsset{compat=1.18}
\usepackage{caption}

\usepackage{microtype}

\usepackage{pifont}
\usepackage{lineno}
\usepackage[normalem]{ulem}
\usepackage[bottom]{footmisc}
\usepackage{float}
\usepackage{lipsum}
\usepackage[linesnumbered,ruled,vlined,nosemicolon]{algorithm2e}
\usepackage{color}

\definecolor{mgreen}{rgb}{0,0.7,0}

\newcommand{\data}{\textsc{PolCA}\xspace}

\usepackage{amsthm,amsmath,amsfonts,bm,xspace}
\usepackage{upgreek}
\usepackage{color}

\def\eqref#1{(\ref{#1})}

\def\1{\bm{1}}

\DeclareMathAlphabet{\mathsfit}{\encodingdefault}{\sfdefault}{m}{sl}
\SetMathAlphabet{\mathsfit}{bold}{\encodingdefault}{\sfdefault}{bx}{n}

\title{Modelling Political Coalition Negotiations Using LLM-based Agents}

\author{Farhad Moghimifar$^1$ \and \textbf{Yuan-Fang Li}$^1$ \\
          \textbf{Robert Thomson}$^2$ \and \textbf{Gholamreza Haffari}$^1$ \\
         $^1$Department of Data Science and AI, Monash University, Australia \\
		 $^2$ School of Social Sciences, Monash University, Australia \\
         \texttt{\{first.lastname\}@monash.edu}}

\begin{document}
\maketitle

\begin{abstract}
Coalition negotiations are a cornerstone of parliamentary democracies, characterised by complex interactions and strategic communications among political parties. Despite its significance, the modelling of these negotiations has remained unexplored with the domain of Natural Language Processing~(NLP), mostly due to lack of proper data. In this paper, we introduce coalition negotiations as a novel NLP task, and model it as a negotiation between large language model-based agents. We introduce a multilingual dataset, \data, comprising manifestos of European political parties and coalition agreements over a number of elections in these countries. This dataset addresses the challenge of the current scope limitations in political negotiation modelling by providing a diverse, real-world basis for simulation. Additionally, we propose a hierarchical Markov decision process designed to simulate the process of coalition negotiation between political parties and predict the outcomes.
We evaluate the performance of state-of-the-art large language models~(LLMs) as agents in handling coalition negotiations, offering insights into their capabilities and paving the way for future advancements in political modelling.
\end{abstract}

\section{Introduction}
\label{sec:intro}
Negotiations for coalition formation are pivotal processes within parliamentary democracies, where multiple political parties cooperate to form a governing alliance based on shared policies and principles~\citep{strom1999keys}. Coalition agreements are crucial for ensuring stable governance, as they outline the collaborative framework through which the coalition will operate, including policy compromises and the distribution of key governmental roles~\citep{kluver2019coalition}. Thus, the negotiation process is inherently complex, involving strategic discussions, bargaining, and compromises to align the diverse interests and policy priorities of the participating parties~\citep{ecker2020coalition}. Simulating the coalition negotiation process offers predictive insights into potential government formations and serves as a valuable tool for researchers, political parties, analysts, and negotiators. Additionally, the simulation provides a valuable learning experience for people participating in democratic process, helping them understand the complexities of multiparty negotiations, coalition management, and the efficient use of resources to achieve favourable outcomes, facilitating the formation of stable and coherent government coalitions~\citep{kluver2019coalition, krauss2023cabinet}.

\begin{figure}
    \centering
    \includegraphics[width=1\linewidth]{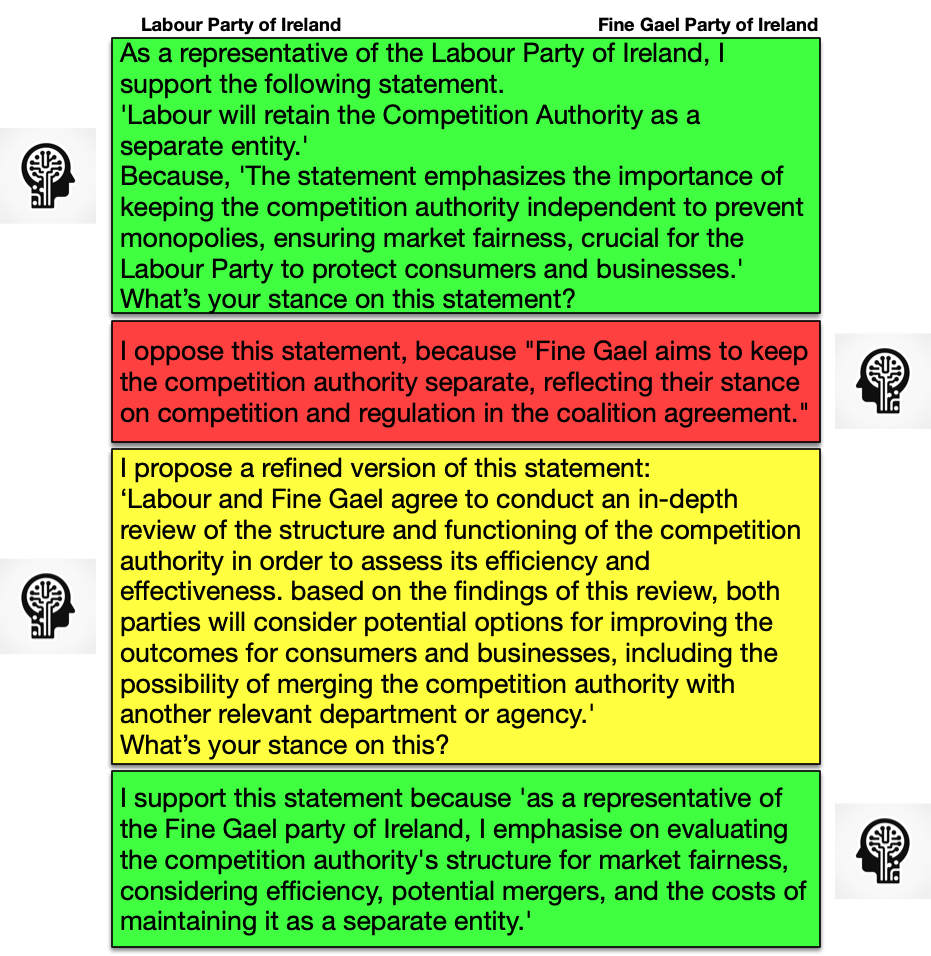}
    \caption{A sample of the negotiation between Fine Gael and Labour party of Ireland, generated by our proposed model. In this sample, the initial statement by the Labour Party was not agreed on by the Fine Gael. The Labour Party proposes a refined version of the initial statement, which eventually Fine Gael supported. In the flow of this negotiation, each parties reasoning for the action taken is also included.}
    \label{fig:enter-label}
\end{figure}

Modelling human negotiations, especially within the realm of political modelling, is a fascinating and challenging problem. However, it has been under-explored, mostly due to a lack of datasets specifically designed for simulating coalition agreements. This gap hinders the development of algorithms capable of accurately simulating the dynamic and complex nature of political coalition formation~\citep{kla2023coalition, muller2024coalition}. Additionally, current approaches in modelling human negotiation processes often exhibit significant shortcomings, primarily due to their limited reasoning scope and depth in understanding the intricacies of human interactions. These models are mostly limited by their focus on surface-level interaction, neglecting the deeper historical and ideological influences that shape negotiation dynamics~\citep{heddaya2023language,chawla2021casino}. They also fall short of fully representing the high-level negotiation process, overlooking critical phases and transitions, and lack the capacity for reflective decision-making based on previous outcomes~\citep{fu2023improving,cheng2023cooper}. This absence of reasoning depth and adaptability in modelling limits their effectiveness in simulating the complex, iterative nature of human negotiations, motivating the need for more sophisticated and realistic approaches.

To tackle the issue of data scarcity in modelling coalition negotiations, we introduce a new, multilingual dataset. This comprehensive dataset comprises political statements from a variety of European countries, enriched with the outcomes of their coalition agreements. The dataset provides a rich foundation for analysing and simulating the complex dynamics of coalition formation and agreement. Additionally, we propose a novel simulation of the coalition negotiation process, where each political party is represented by a Large Language Model~(LLM)-based agent. These agents operates under a hierarchical Markov Decision Process~(MDP) formulation, designed to simulate the intricacies of the negotiation process. This hierarchical approach allows the agents to capture different aspects of the negotiations, from initial discussions to final agreements. As the negotiation progresses, these LLM-agents learn and update their policies, adapting their strategies based on the unfolding interactions and decisions. This method enables a dynamic simulation of coalition negotiations, where agents iteratively refine their approaches to reach a consensus, mirroring the adaptability required in real-world political negotiations.

We conduct a comprehensive evaluation of the performance of state-of-the-art large language models in executing coalition negotiations. This evaluation provides critical insights into the capabilities and limitations of current language models in accurately simulating complex political processes, paving the way for future advancements in the field. Our experimental results show that while the complexity of the coalition negotiations makes predicting the outcomes difficult, our proposed model can achieve reasonable results in delivering this task.

\section{Related Works}
\label{sec:lit_rev}

\paragraph{Political Science in NLP.}
Traditionally, political science has used computational approaches to study a variety of areas, such as policy making~\citep{wyner2010policy} and voting analysis~\citep{ansari2020analysis,budhwar2018predicting}. More advanced computational techniques have been applied for topic modelling~\citep{parthasarathy2019deliberative}, political text classification~\citep{chang2020using} and sentiment analysis~\citep{soroka2015bad}. Furthermore, there has been an interest from computer science community to apply computational techniques to political science problems. For instance, \citet{iyyer2014political,liu2022politics,baly2020we} looked into ideology detection, and \citet{zhou2019fake,wang2023attacking} tackled the task of fake news detection. With recent advances in LLMs, there's been a surge in using such models to replace manual processes in political science~\citep{alizadeh2023open}. By leveraging the reasoning capabilities of these models that are trained on large amount of data, recent research have shown that LLMs can outperform human annotation in detecting the political affiliation of social media users~\citep{tornberg2023chatgpt} and topic, sentiment, stance, and frame identification~\citep{gilardi2023chatgpt}. More aligned with our task in this paper, there has been some efforts in applying NLP techniques to understand the content of political party manifestos~\citep{orellana2023using,bonisch2023bundestag}. While the process of coalition negotiation has been extensively researched in the political science community~\citep{guinaudeau2020electoral, adhikari2024fulfilment, costello2008election}, to the best of our knowledge this work is the first effort from computational perspective.
\paragraph{Agent-based Modelling.} In Modelling complex systems, agent-based approaches have been widely used to capture interactions and emergent behaviours in the environment~\citep{macal2005tutorial,salgado2013agent}, where recent integration of LLMs into such approaches has shown a promising direction for exploration~\citep{zhao2023survey}. LLM's capabilities such as human-like planning and scheduling, interactive responding~\citep{gravitas2023autogpt,park2023generative}, and not requiring pre-defined instructions~\citep{wang2023survey} have motivated deploying LLMs as the core engine of simulating complex systems. Specifically, in modelling human interactions \citet{park2023generative} implemented a sandbox of social life, where they showed the capability of LLMs in navigating through a daily life and delivering variety of tasks. In more cooperative scenarios, \citet{liang2023encouraging,saha2023can} demonstrated the effectiveness of leveraging multiple agents arguments in solving a task. Some efforts have applied LLM agent-based simulations to negotiation scenarios, which are closer to our proposed task in this research. \citet{fu2023improving} used a LLM to provide feedback to other LLMs who are performing item purchase negotiation. \citet{bianchi2024well, abdelnabi2023llm} applied LLM-based multi-round negotiation systems to various scenarios and showed their capability in reaching agreements. Contrary, to these works, the task of coalition negotiation requires capturing long-term complexities in multiple political parties preferences and taking consequential actions, which has been overlooked in other related works by simplifying the simulation setups.

\section{The Coalition Negotiation Task}
\label{sec:data}

The task of coalition negotiation entails predicting the outcome of complex negotiations between political parties aiming to form a coalition government. The starting point for this predictive task is the set of statements (i.e.\ sentences) within the manifesto of each party involved in the negotiations. A manifesto is a public declaration of policies and aims, often issued by political parties to outline their intended programs and core principles. These manifestos contain key policy positions and intentions that the parties wish to pursue once in government. The negotiation process involves parties coming together to discuss, compromise, and eventually agree on a shared policy agenda that is reflected in the final coalition agreement. A coalition agreement is a formal accord reached between political parties, outlining the policies and principles they agree to implement jointly in a coalition government. The success of the negotiation process is contingent upon how well the diverse policy positions, as represented by parties' manifestos, can be reconciled into a cohesive plan, taking on the form of the final coalition agreement, that all parties can support.

In this section, we formally formulate the coalition negotiation task (\S\ref{sec:prob}), followed by a presentation of the construction process of our dataset, describing the constituent datasets (\S\ref{sec:datasets}) and the annotation process (\S\ref{sec:annotation}). 

\subsection{Problem Formulation}\label{sec:prob}
For the simplicity of modelling, we assume the negotiation occurs between two parties only. However, our formulation and method can be readily extended to multi-party negotiations.

Let $\bm{P}=\{p,q\}$ represent two political parties in a given country. The manifesto of party $p$ is represented as a set of statements $M_{p} = \{m^1_{p},m^2_{p},\ldots,m^n_{p}\}$, where $|M_{p}|=n$. Let $C_{p,q} = \{c_{p,q}^1, \dots, c_{p,q}^m\}$ represent the final coalition agreement. For each manifesto sentence $m^k_{p}$ of party $p$, we define  the \emph{negotiation task} as determining whether $m^k_{p}$ (partly) exists in the final coalition agreement $C_{p,q}$ or not. We consider this  as a negotiation task, because by nature, the process of negotiation and history of the negotiation potentially affects the decision about a statement. To characterise the outcome, we adopt the labels defined by \citet{guinaudeau2020electoral}\footnote{This dataset is not publicly available.}: $\xi=\{\emph{0: not included}, \emph{1: partly included}, \emph{2: included}\}$.

Formally, for each party $p$, we aim to learn a function $f_p: M_{p} \rightarrow \xi$ that predicts, for each manifesto statement $m_{p}^k\in M_{p}$, whether it is (partly) included in the final coalition agreement $C_{p,q}$. 

To evaluate the accuracy of models tasked with predicting the outcomes of these negotiations, gold labels are required based on whether each statement from the initial party manifestos is included in the final coalition agreement. This method of evaluation allows for a precise measurement of a model's performance in simulating the negotiation process. By comparing the model's predictions to the gold labels, we can assess the effectiveness of different approaches in accurately capturing the dynamics and outcomes of coalition negotiations.

\begin{table*}[ht]
    \centering
    \resizebox{\linewidth}{!}{
	       \begin{tabular}{l|c c c c c c c c c c c c}
            \hline
	           & \multicolumn{2}{c}{\textbf{Austria 2013}} & \multicolumn{2}{c}{\textbf{Germany 2013}} & \multicolumn{2}{c}{\textbf{Iceland 2013}} &
                 \multicolumn{2}{c}{\textbf{Ireland 2011}} & \multicolumn{2}{c}{\textbf{Netherlands 2012}} & \multicolumn{2}{c}{\textbf{Portugal 2012}}\\
            \cline{2-13}
            Political Party & SDP & APP & 
                              CDU & SDP & 
                              IP & PP &
                              FG & LP &
                              PPFD & LP &
                              SDP & CDS-PP\\
            \hline
            \#statements included & 103 & 219 & 594 & 559 & 7 & 12 & 332 & 330 & 121 & 159 & 55 & 3\\
            \#statements partly included & 405 & 615 & 1,354 & 1,517 & 54 & 62 & 710 & 791 & 782 & 1,147 & 843 & 219 \\
            \#statements not included & 217 & 328 & 625 & 822 & 58 & 43 & 458 & 314 & 810 & 1,452 & 1,795 & 843\\
            \#total statements & 725 & 1,162 & 2,574 & 2,898 & 119 & 117 & 1,500 & 1,435 & 1,713 & 2,758 & 2,693 & 1,065\\
            \hline
	       \end{tabular}}
    \caption{Summary statistics of the dataset for six  countries and their political parties which formed a coalition agreement in the respective year. It shows the total number of statements in the manifesto of each party, as well as the number of the statements that were included, partly included and not included in the final coalition agreement.}
    \label{tab:dataStat}
\end{table*}

\subsection{\data: A Dataset for Political Coalition Agreement}\label{sec:datasets}
In order to address the data scarcity problem in this area, we introduce a new dataset, \data, which is compiled by analysing and annotating the content of  two existing datasets in political science, as explained in the following. 

\paragraph{The Coalition Agreement Dataset.}
To model the outcome of the negotiation process, we have collected and processed data from the COALITIONAGREE~\citep{DVN/XM5A08_2023} dataset. This dataset compiles finalised coalition agreements from various governments, capturing the outcomes of intricate negotiation processes. These agreements are pivotal as they represent the consensus and compromises made by the parties involved, reflecting the practical application of their manifesto commitments in a collaborative governance context. To ensure comprehensive and accurate modelling, the dataset has been curated from official government sources and political archives, encompassing a diverse range of political contexts and coalitions. It has been processed to align with the initial manifesto data, facilitating an integration in the modelling process. 

\paragraph{The Manifesto Project Dataset.}
Manifestos are critical in setting the tone and direction of the negotiations, as they reflect the commitments and priorities that parties aim to uphold and negotiate upon when forming a coalition government. In the context of collecting data for modelling coalition negotiations, we utilise the data from the ``Manifesto Project''~\citep{lehmann2023manifesto}\footnote{\url{https://manifesto-project.wzb.eu}} as the foundational element. This project offers a comprehensive database of 1,000 political party manifestos from 50 countries, providing a rich source of information on party positions, policy preferences, and ideological orientations. By leveraging this data, we can simulate the starting point of coalition negotiations, where parties' manifestos serve as the basis for initial discussions and bargaining.

\begin{figure}[t]
    \centering
    \includegraphics[width=0.75\linewidth]{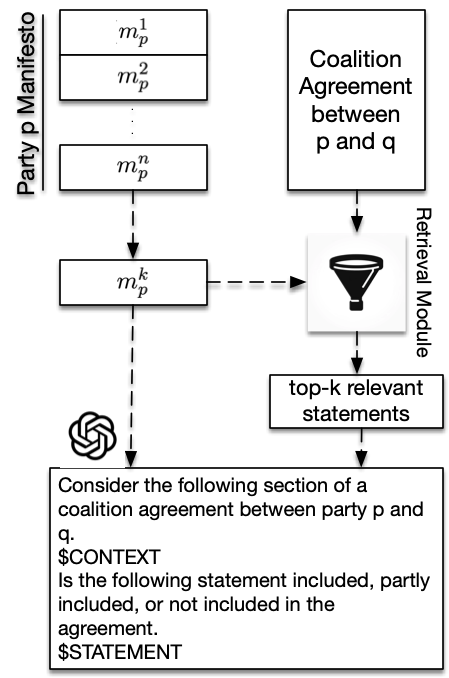}
    \caption{Graphical representation of our data collection pipeline. A statement from a manifesto of party $p$ is selected, $m_p^k$. We use a \emph{Retrieval Module} to scope down the context to relevant statements from the coalition agreement dataset. The statement and set of relevant context is fed to GPT-4 to collect the annotation.}
    \label{fig:enter-label}
\end{figure}

\begin{figure*}[!ht]
  \subfloat[\centering \tiny {Austria 2013 - Left: Social Democratic Party  - Right: Austrian Poeple’s Party}]{
	\begin{minipage}[c][1\width]{
	   0.16\textwidth}
	   \centering
	   \includegraphics[width=1\textwidth]{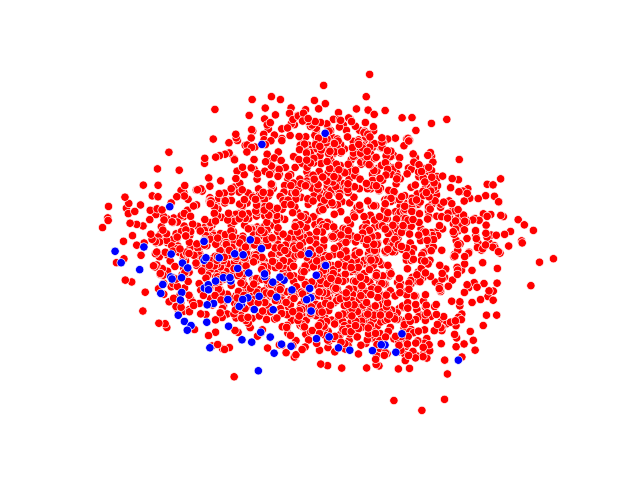}
	\end{minipage}
	\begin{minipage}[c][1\width]{
	   0.16\textwidth}
	   \centering
	   \includegraphics[width=1\textwidth]{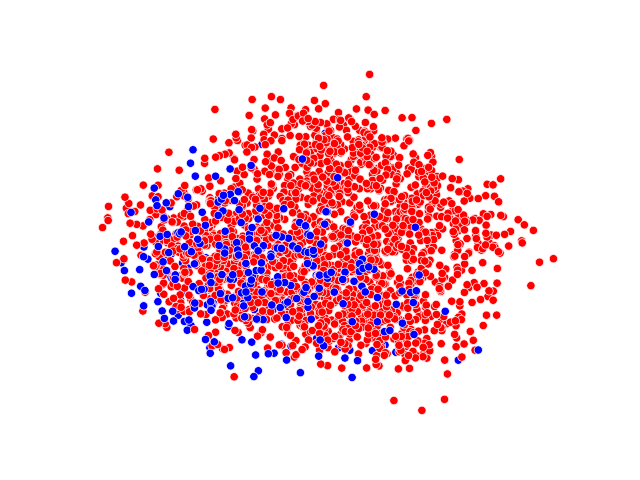}
	\end{minipage}}
 \hfill
 \subfloat[\centering \tiny Germany 2013 - Left: Christian Democratic Union  - Right: Social Democratic Party]{
	\begin{minipage}[c][1\width]{
	   0.16\textwidth}
	   \centering
	   \includegraphics[width=1\textwidth]{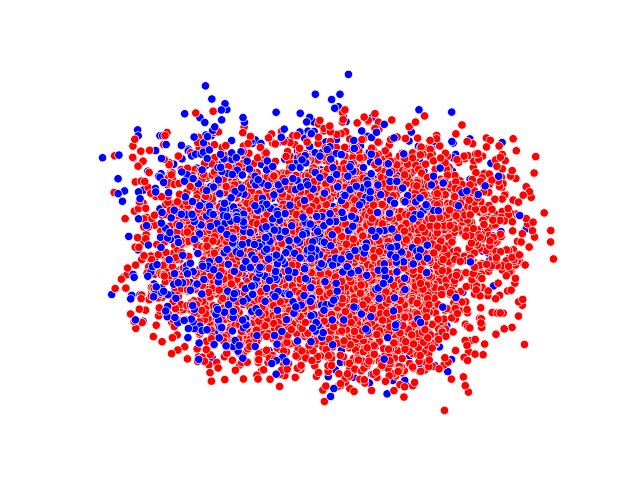}
	\end{minipage}
	\begin{minipage}[c][1\width]{
	   0.16\textwidth}
	   \centering
	   \includegraphics[width=1\textwidth]{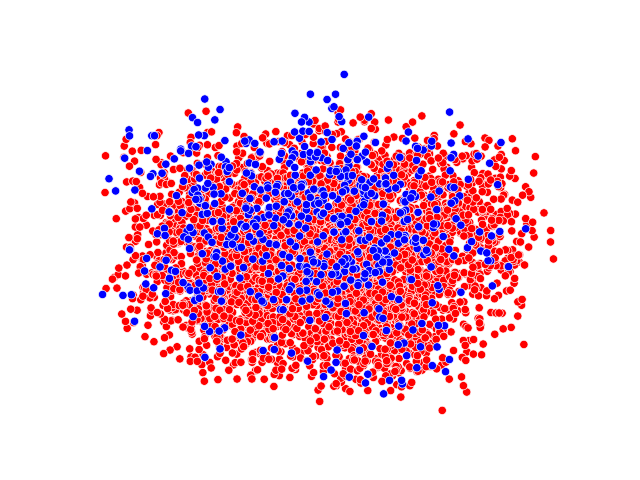}
	\end{minipage}}
 \hfill 
 \subfloat[\centering \tiny Iceland 2013 - Left: Independence Party  - Right: Progressive Party]{
	\begin{minipage}[c][1\width]{
	   0.16\textwidth}
	   \centering
	   \includegraphics[width=1\textwidth]{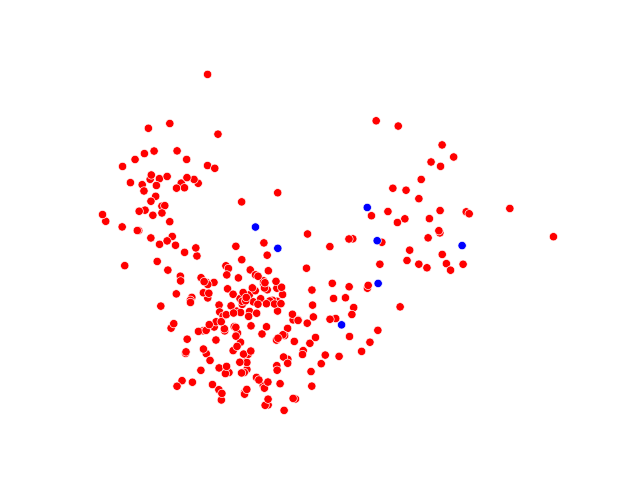}
	\end{minipage}
	\begin{minipage}[c][1\width]{
	   0.16\textwidth}
	   \centering
	   \includegraphics[width=1\textwidth]{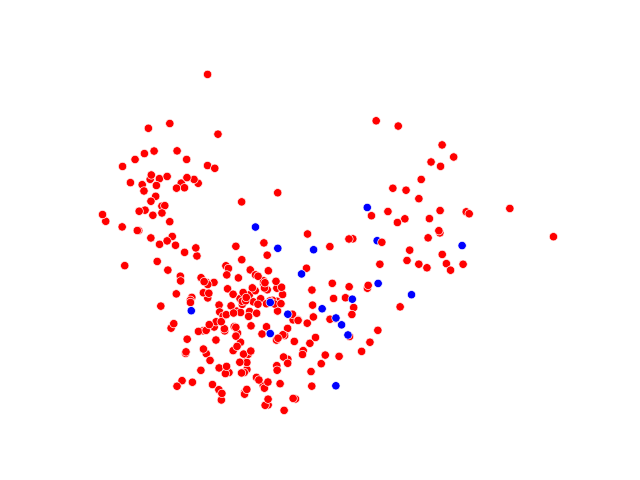}
	\end{minipage}}
    \hfill 
 \subfloat[\centering \tiny Ireland 2011  - 
            Left: Fine Gael  - 
            Right: Labour Party]{
	\begin{minipage}[c][1\width]{
	   0.16\textwidth}
	   \centering
	   \includegraphics[width=1\textwidth]{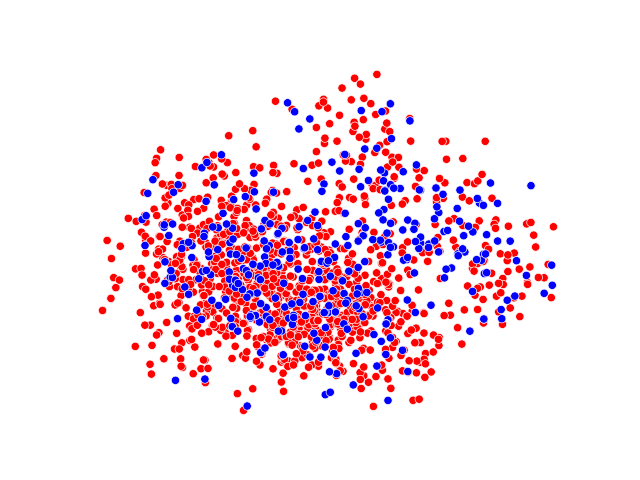}
	\end{minipage}
	\begin{minipage}[c][1\width]{
	   0.16\textwidth}
	   \centering
	   \includegraphics[width=1\textwidth]{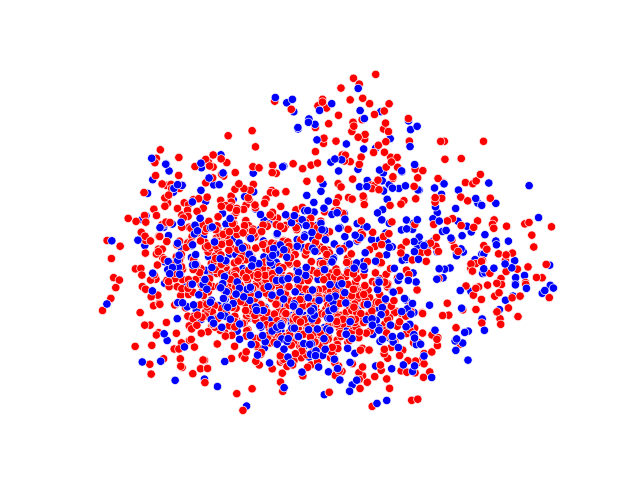}
	\end{minipage}}
 \hfill 
 \subfloat[\centering \tiny Netherlands 2012  - 
            Left: People’s Party for Freedom and Democracy  - 
            Right: Labour Party]{
	\begin{minipage}[c][1\width]{
	   0.16\textwidth}
	   \centering
	   \includegraphics[width=1\textwidth]{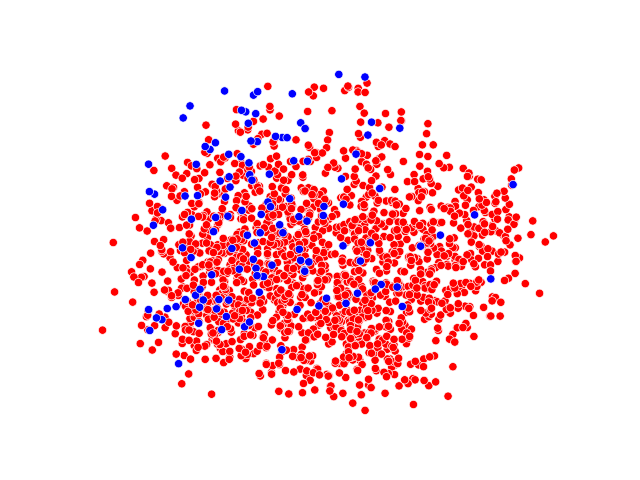}
	\end{minipage}
	\begin{minipage}[c][1\width]{
	   0.16\textwidth}
	   \centering
	   \includegraphics[width=1\textwidth]{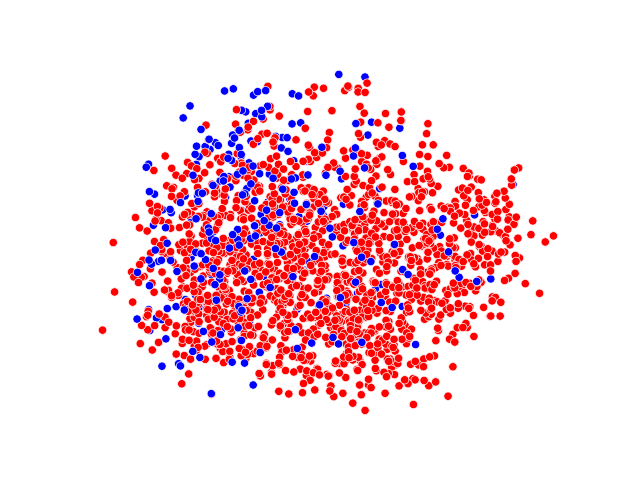}
	\end{minipage}}
 \hfill 
 \subfloat[\centering \tiny Portugal 2012  -  Left: Social Democratic Party  -  Right: CDS – People’s Party]{
	\begin{minipage}[c][1\width]{
	   0.16\textwidth}
	   \centering
	   \includegraphics[width=1\textwidth]{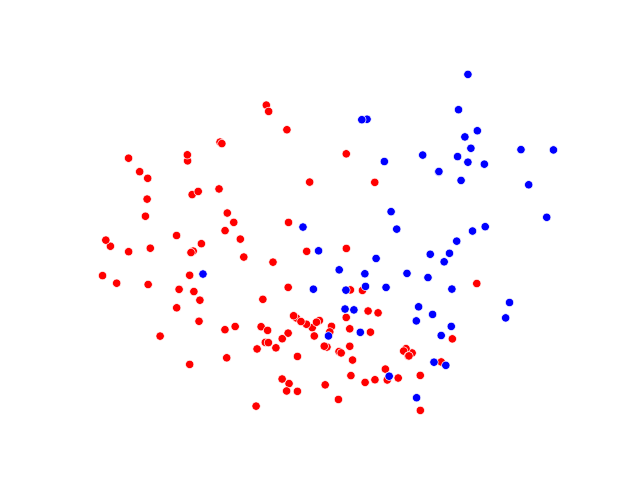}
	\end{minipage}
	\begin{minipage}[c][1\width]{
	   0.16\textwidth}
	   \centering
	   \includegraphics[width=1\textwidth]{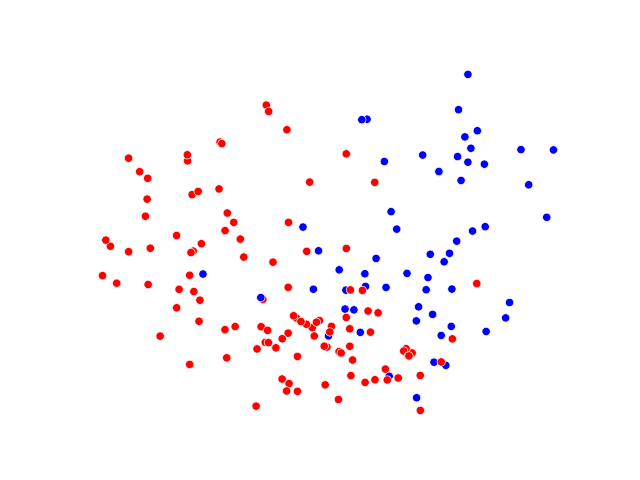}
	\end{minipage}}
    \caption{The distribution of the statements that are included in the final coalition agreement~(blue) over the statements of the coalition agreement~(red) per party per country. We have used PCA~\citep{wold1987principal} as dimensionality reduction method to visualise this information.}
    \label{fig:dist}

\end{figure*}

\subsection{Collecting Annotations}\label{sec:annotation}
To connect the above-mentioned datasets, we have developed an automated annotation process aimed at determining which statements from the manifestos are incorporated into the final coalition agreement. 
We aim to obtain, for each statement in $m^i_p\in M_p$, the most likely label from $\xi$:
\begin{equation}
    f_p(m^i_p) = \text{argmax}_{e\in\xi}P_{\theta_{class}}(e|m^i_p,C_{pq})
\end{equation}
For our annotation purposes, we have used OpenAI's GPT-4~\citep{achiam2023gpt} as $P_{\theta_{class}}$ and a retrieval model based on FAISS~\citep{johnson2019billion}\footnote{\texttt{https://github.com/facebookresearch/faiss}} to provide the top-$k$ relevant statements from COALITIONAGREE as the context. Relevance is calculated as the Cosine similarity between the encoded representation of the statements by using OpenAI's \texttt{text-embedding-ada-002} model. Given the identified top-$k$ statements, we prompt GPT-4 to determine whether statement $m^i_p$ is not included, partly included, or included in the set of statements. 
Figure \ref{fig:enter-label} shows our data annotation pipeline.

Table \ref{tab:dataStat} shows a summary of the statistics of our collected data. We have selected six countries, Austria, Germany, Iceland, Ireland, Netherlands, and Portugal between 2011 to 2013. During this period, the democratic system in each of these countries faced a coalition negotiation. The following coalition governments were formed: in Austria, 2013, Social Democratic Party~(SDP) and Austrian Poeple's Party~(APP); in Germany, 2013, Christian Democratic Union~(CDU) and Social Democratic Party~(SDP); in Iceland, 2013, Independence Party~(IP) and Progressive Party~(PP); in Ireland, 2011, Labour Party(LP) and Fine Gael~(FG); in Netherlands, 2012, People’s Party for Freedom and Democracy~(PPFD) and Labour Party~(LP); and in Portugal, 2011, Social Democratic Party~(SDP) and CDS – People’s Party~(CDS-PP). Table \ref{tab:dataStat} shows the total number of statements in each party's manifesto, and the number of the statements that were included, partly included, and not included in the final coalition agreement. Figure \ref{fig:dist} visualises the distribution of the statements from each party   included in the final agreement~(blue dots) over the statements of the coalition agreement~(red dots).

\section{LLM-based Agents for Political Coalition Negotiation}
\label{sec:approach}

We propose a \emph{Hierarchical Markov Decision Process} approach to simulate the political coalition negotiation process, and identify whether each statement should be included in the final agreement. We explain our approach in the rest of this section. 


\begin{figure}[t]
    \centering
    \includegraphics[width = \linewidth]{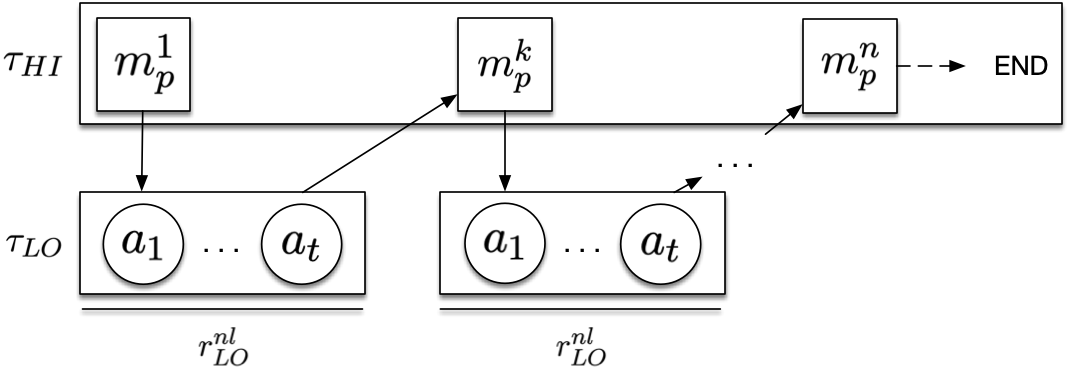}
    \caption{Our proposed hierarchical MDP framework. At the higher level, using $\sigma_i^{HI}$, the agent chooses which statement to negotiate. At the lower level, based on $\sigma_i^{LO}$ the agent takes action about the statement. The set of actions at each level generates a trajectory $\tau_i^{LO}$ and $\tau_i^{HI}$. Using the reward function $r^{nl}_{LO}$ and $r^{nl}_{HI}$, the agent updates the respective policy.}
    \label{fig:HRL}
\end{figure}

We formulate the negotiation between the agents (i.e.\ political parties)  as {``normal-form game''} $(\Pi, \bm{U}, n)$, where $n$ denotes the number of the players, $\Pi = \{\Pi_1, \dots, \Pi_n\}$ is the set of policies of strategies corresponding to each player, and $\bm{U}$ is the payoff for each joint policy played by all players. In this game, each player tries to maximise their own expected payoff by choosing a policy $\sigma_i \in \Pi_i$, where the quality of $\sigma_i$ is affected by other players' strategies. Our aim in this paper is not to find the optimal policies, instead, it is to learn  (approximately) the underlying policies used by the political parties in their negotiations based on the \data dataset. 

We model the coalition negotiation as a two-level hierarchical Markov decision process. At the higher ($HI$) level, an agent $p$ chooses statement $m_p^i$ from $\bm{M} = \{m_p^1,\dots, m_p^n\}$. At the lower ($LO$) level, the agent chooses action $a_k$ from $\bm{A}=\{a_1, \dots, a_t\}$ to negotiate the chosen statement at the higher level. An agent's choice at both levels depends on the observed state $s\in S$. The horizon for $HI$ is the total number of the statements to be discussed during the negotiation, and the horizon at level $LO$ is the a predefined number of negotiation rounds over a statement. We define a termination function $\beta : S \rightarrow \{True, False\}$ to determine the termination of execution of the policy at the $LO$ level. The agent iteratively delivers choices at both levels until a termination condition (eg the max horizon length at each level) is reached.

The goal here is for an agent $p_i$ to learn a policy at the high level $\sigma_i^{HI}: S \rightarrow M$, while learning a policy at the low level $\sigma_i^{LO}: S \rightarrow A$, where the payoff $\bm{U}$ is maximised~($\sigma_i^{HI} \in \Pi_i$ and $\sigma_i^{LO} \in \Pi_i$). The execution of each policy at low level, $\sigma_i^{LO}$, generates a trajectory $\tau_i^{LO} = (s_1, a_1, \dots, s_H, a_H, s_{H+1})$ and combined with the execution of the high level policy $\sigma_i^{HI}$, the overall trajectory then is presented as: $\tau_i^{HI} = (s_1, a_1, \tau_{i_1}^{LO}, s_2, a_2, \tau_{i_{2}}^{LO}, \dots)$.

At each iteration, the policies at both levels are updated using the input from the reward functions $R_{LO}$, $R_{HI}$, and a $\texttt{Critique}$ function. At the low level, the reward function $R_{LO}: S \times A \rightarrow \{True, False\}$ indicates whether the agent has been successful in negotiating the statement. At the high level, the reward function $R_{HI}: S \times M \rightarrow \mathbb{N}$, indicates how similar the outcome of the negotiation is to the ground truth. The function $\texttt{Critique}$ provides detailed analysis and assessment of the agent in the previous step of the negotiation, based on the trajectory and the rewards, at the low and high levels. The process of our proposed HMDP is summarised in Algorithm~\ref{alg:hmdp}.


{
\begin{algorithm}
\caption{Hierarchical MDP for political negotiations}
\label{alg:hmdp}
\small
\For{$h_{HI} = 1\ldots H_{HI}$}{
    Observe state $s$ and choose action (statement)$m\leftarrow\pi_{HI}(s)$\;
    \For{$h_{LO}=1\ldots H_{LO}$}{
        Observe state $s$\;
        \lIf{$\beta(s)$}{
            \textbf{Break}
        }
        Choose action $a\leftarrow \pi_{LO}(s)$\;
        $\tau_{LO}=(s_1,a_1,\dots,s_H,a_H,s_{H+1})$\;
        $r^{nl}_{LO} = R_{LO}(\tau_{LO})$\;
        Update $\sigma_i^{LO} \leftarrow \texttt{Critique}(\tau_{LO}, r^{nl}_{LO})$\;
    }
    $\tau_{HI} = (s_1,p_1,s_2,p_2,\dots)$\;
    $r^{nl}_{HI} = R_{HI}(\tau_{HI})$\;
    Update $\sigma_i^{HI} \leftarrow \texttt{Critique}\;(\tau_{HI}, r^{nl}_{HI})$\;
}
\end{algorithm}
}

\section{Experiment}
\label{sec:experiment}
In this section we report the performance of our proposed Hierarchical Markov Decision Process~(\textsc{hMDP}) in simulating coalition negotiations in various setup. By comparing the output of the simulation with the outcome of the real world scenarios, we measure how realistic the proposed controlled simulation is. Moreover, we initialise the agents with different large language models (LLMs) to leverage their strong understanding of commonsense and political knowledge as well as strong reasoning capabilities. In this way, we study the effectiveness of LLMs in addressing this task as well. 

\paragraph{Model Engines.}
We consider different LLMs as the backbone of our approach. The main criteria for our setup is to have a chat-based LLM. Therefore, we have selected OpenAI's \texttt{gpt-3.5-turbo}, \texttt{LLaMa-7b-chat}, and \texttt{LLaMa-13b-chat}~\citep{touvron2023llama}.

\paragraph{Implementation Details.}
In our experiments, we have considered a set of pre-defined actions for the agents. These actions include \emph{support}: the agent supports the statement to be included in the final agreement, \emph{oppose}: the agent does not support the statement to be included in the final agreement, \emph{refine}: the agent proposes a new refined version of the statement, and \emph{compromise}: the agent changes its own stance on a previously opposed statement from the other party, to use this as a bargaining chip for negotiating over its own current statement. Negotiation over each statement happens for maximum of three rounds~(i.e.\ $H_{LO} = 3$), unless an agreement is reached for $h_{LO} < H_{LO}$. For $R_{LO}$ we have used natural language representation of the outcome of the previous round of the negotiation, where updating the agent whether previous step was successful or not. $R_{HI}$ is determined by the overall accuracy of the model in comparison to the ground truth. For all of the experiments we set the LLM temperature to 0.5, and for reproduciblity of the results we fix the random seed at 111. The details of the prompts used in this paper can be found in Appendix~\ref{sec:prompts}.
\begin{table*}[ht]
    \centering
    \resizebox{0.98\textwidth}{!}{
    \begin{tabular}{l l c c c c c c c c c c c c c c c}
        \hline
         &  & \multicolumn{2}{c}{\textbf{Austria 2013}} & \multicolumn{2}{c}{\textbf{Germany 2013}} & \multicolumn{2}{c}{\textbf{Iceland 2013}} & \multicolumn{2}{c}{\textbf{Ireland 2011}} & \multicolumn{2}{c}{\textbf{Netherlands 2012}} & \multicolumn{2}{c}{\textbf{Portugal 2011}}  \\
        \textbf{Policy}       & \textbf{Engine}     & SDP    & APP  & CDU   & SDP   & IP & PP & FG & LP & PPFD & LP & SDP & CDS-PP \\
        \hline
        \multirow{3}{*}{\textsc{hMDP}}   & {\texttt{gpt-3.5}}   & 48.29 & 53.44 & 44.66 & 44.66 & 46.56 & 50.14 & 43.77 & 43.50 & 48.16 & 48.51 & 49.65 & 36.46 \\
                              & {\texttt{llama-13b}} & 26.54 & 31.72 & 29.9  & 40.80 & 35.43 & 38.02 & 38.60 & 33.13 & 43.44 & 34.98 & 41.49 & 41.71\\
                              & {\texttt{llama-7b}}  & 31.55 & 34.12 & 33.28 & 37.62 & 32.47 & 33.81 & 30.03  & 27.98 & 30.73 & 30.73 & 26.28 & 29.24 \\
        \hline
        \multirow{3}{*}{\textsc{hMDP}-\texttt{LO}}   & {\texttt{gpt-3.5}}   & 47.96 & 44.58 & 50.07 & 54.82 & 39.97 & 52.93 & 48.11 & 51.42 & 40.80 & 46.77 & 41.89 & 33.26\\
                              & {\texttt{llama-13b}} & 26.1  & 33.6  & 30.03 & 29.74 & 29.54 & 26.57 & 38.11 & 32.53 & 29.22 & 29.73 & 24.14 & 31.59 \\
                              & {\texttt{llama-7b}}  & 31.26 & 28.38 & 15.82 & 25.05 & 14.37 & 14.37 & 15.63 & 24.23 & 21.95 & 36.93 & 20.16 & 26.26 \\
        \hline
        \multirow{3}{*}{\textsc{hMDP}-\texttt{Base}} & {\texttt{gpt-3.5}}   & 49.54 & 42.78 & 46.64 & 51.88 & 43.85 & 47.76 & 51.01 & 50.20 & 40.00 & 43.52 & 39.71 & 31.67\\
                              & {\texttt{llama-13b}} & 29.32 & 30.85 &  25.92 & 39.22  & 24.65 & 28.95 & 42.17 & 32.53 & 29.65 & 29.71 & 12.2  & 25.95\\
                              & {\texttt{llama-7b}}  & 23.44 & 23.45 & 13.56 & 26.13 & 21.08 & 19.65 & 27.74 & 23.5  & 24.45 & 27.75 & 10.97 & 10.97\\
        \hline
    \end{tabular}}
    \caption{The macro F1-score of the approaches in modelling the result of coalition negotiations in different European countries in their respective languages.}
    \label{tab:ablation}
\end{table*}

\paragraph{Metric.}
We select macro F1 as the metric mostly due to the classification nature of the task. Considering the imbalanced distribution of the classes in the dataset, this metric ensures an unbiased evaluation across all categories.

\paragraph{Results.}
Table~\ref{tab:ablation} summarises the performance of our proposed model, \textsc{hMDP}, with different LLMs as the backbone engine. As can be seen, predicting the outcome of coalition negotiations 
remains a very challenging task for highly capable LLMs.
This is due to several factors, including the complexity of the language used in the political manifestos and coalition agreements, as well as inherent limitations of the LLMs' knowledge and understanding of the specific negotiation processes prevalent in each country. The LLMs' performance is significantly influenced by their ability to parse and interpret the nuances of political language and strategy, which can vary widely from one national context to another. Additionally, the internal models and data on which these LLMs are trained may offer differing degrees of familiarity with the political landscapes of various countries, affecting their ability to accurately simulate negotiation outcomes.

Within the simulations conducted for each country, we observe in Table~\ref{tab:ablation} that agents representing different political parties exhibited distinct behaviours during the negotiation process. For instance, in Portugal 2011, a 13 point gap between our model's prediction can be seen. This diversity in behaviour primarily stems from our experimental setup, in which each agent is programmed to negotiate with the objective of maximising its own benefits. As a result, the negotiation dynamics become a complex interplay of strategies, with each agent seeking to optimise its own policy preferences and strategic goals. These differences in behaviour underscore the competitive nature of political negotiations, where parties are driven by the desire to secure the most favourable terms for their constituencies. The observed variations highlight the nuanced and individualised approaches parties take in negotiations, influenced by their unique priorities and the strategic considerations of maximising political gains while contributing to coalition formation.

\begin{table*}[ht]
    \centering
    \resizebox{0.98\textwidth}{!}{
    \begin{tabular}{l l c c c c c c c c c c c c c c c}
        \hline
         &  & \multicolumn{2}{c}{\textbf{Austria 2013}} & \multicolumn{2}{c}{\textbf{Germany 2013}} & \multicolumn{2}{c}{\textbf{Iceland 2013}} & \multicolumn{2}{c}{\textbf{Ireland 2011}} & \multicolumn{2}{c}{\textbf{Netherlands 2012}} & \multicolumn{2}{c}{\textbf{Portugal 2011}}  \\
        \textbf{Method} & \textbf{Engine} & SDP & APP & CDU & SDP & IP & PP & FG & LP & PPFD & LP & SDP & CDS-PP \\
        \hline
        \textsc{hMDP} & \texttt{gpt-3.5-turbo} & 48.29 & 53.44 & 44.66 & 44.66 & 46.56 & 50.14 & 43.77 & 43.50 & 48.16 & 48.51 & 49.65 & 36.46 \\
        \emph{classifier}       & \texttt{gpt-3.5-turbo} & 49.93 & 47.08 & 45.83 & 51.58 & 48.48 & 47.29 & 41.35 & 40.89 & 43.73 & 48.04 & 44.05 & 39.60 \\
        \emph{Open-Neg}          & \texttt{gpt-3.5-turbo} &  46.96 & 45.11 &  43.47 & 44.66 & 48.48 & 47.29 & 42.43 & 41.52 & 48.16 & 48.51 & 49.48 & 49.92 \\
        \hline
    \end{tabular}}
    \caption{The accuracy of the approaches in modelling the result of coalition negotiations in different European countries in their respective languages.}
    \label{tab:sim_ablation}
\end{table*}

Furthermore, by analysing the performance of our model with different backbone engines, it can be observed that \texttt{gpt-3.5-turbo} consistently outperforms both \texttt{LLaMa-13b-chat} and \texttt{LLaMa-7b-chat} models. This superiority in performance can be attributed to the correlation between the number of parameters in these models and their respective abilities to accurately simulate the coalition negotiation process. The \texttt{gpt-3.5-turbo} model, with its higher parameter count, demonstrates better understanding and reasoning capabilities, resulting in more realistic simulations. 

\subsection{Ablation Study}

\paragraph{Policy Learning.}
To measure the effect of our policy learning method, we report the performance of our proposed approach in two variants. Firstly, we remove the higher-level policy, and resort to lower-level policy only for negotiating over the statement, denoted by \textsc{hMDP}-\texttt{LO}. In this setting, each agent randomly chooses the statement to be discussed next. Secondly, we remove both higher and lower level policy during the negotiation, where the agents act with local information related to the statement only, denoted as \textsc{hMDP}-\texttt{Base}.

Table \ref{tab:ablation} summarises the performance of our model in these two settings. As it can be seen, in comparison to \textsc{hMDP} policy setting, in the majority of countries and political parties analysed, the models employing both higher-level and lower-level policies outperform those relying on lower-level policies alone or without any policy learning modules. This finding highlights the effectiveness and importance of utilising the hierarchical approach to policy learning in negotiation simulations. Implementing a nuanced policy framework, which includes both overarching strategies and detailed action plans, enhances the models' ability to navigate complex negotiation scenarios, leading to more accurate outcomes. 

During error analysis we observe that the performance of models employing both higher and lower level policies was occasionally negatively impacted by LLMs' hallucinations in generating outputs that align with the format required for the simulation. An example of such a discrepancy is the generation of statement IDs that do not match any of the IDs from the provided list, introducing errors into the simulation process. Despite these challenges, using lower-level policies (\textsc{hMDP}-\texttt{LO}) alone still yielded better results than the base setting (\textsc{hMDP}-\texttt{Base}), highlighting the value of detailed policy frameworks in enhancing simulation fidelity. Addressing the identified output format issues, such as ensuring the accuracy generated output, could further improve the overall results.

\paragraph{Simulation.}
We evaluate the effectiveness of performing this task in a controlled simulated situation by comparing the performance of our model~(\textsc{hMDP}) with two baselines. Firstly, we treat this task as a classification task without any negotiation between the agents. For this purpose, for each statement in a manifesto of a party, we prompt an LLM with the statement and instruct it to identify whether the statement should be in the final agreement or not, denoted as \emph{Classifier}. Secondly, we perform the task in an open negotiation manner, where the agents are not instructed with a set of predefined actions, and after three rounds of negotiation, we ask an external LLM to identify whether the agents have come to an agreement over whether to include the statement in the final agreement or not, denoted as \emph{OpenNeg}. In both settings, we use \texttt{gpt-3.5-turbo} as the backbone engine.

Table \ref{tab:sim_ablation} summarises the results. As can be seen, employing a simple classifier for simulating coalition negotiations yields lower performance, mostly due to the complexity of the negotiation process where the outcome of each step can significantly influence future steps. This complexity is not properly captured by simplistic models, which fail to account for the iterative and dynamic nature of negotiations. Additionally, the \emph{OpenNeg} model, designed to test open-ended negotiation simulations, also demonstrates lower performance, suggesting that the general capabilities of broadly trained LLMs might not be fully equipped to handle the specific demands and nuances of negotiation simulations. This issue could potentially be addressed by developing customised or specialised LLMs tailored specifically for simulating negotiation processes.

\section{Conclusion}
\label{sec:conclusion}
We presented a publicly-available multi-lingual dataset, \data, for predicting the outcome coalition negotiation in six European Countries. Specifically, we focused on annotating the manifestos of political parties in these countries, by comparing their content with the coalition agreements. \data supports a novel task in NLP~(Section~\ref{sec:data}), where by proposing a new hierarchical Markov decision process, we scratched the surface in tackling this challenging task. Through empirical studies, we showed the capability of different Large Language Models, as a standalone model, or incorporated in our proposed model, in addressing this task. We observed that the model performance is affected by the number of parameters of the model, language of the data, and their training process.


\section{Ethics Statement}
As we use existing LLMs, without further tuning, our approaches do not introduce any new ethical concerns during automatic data annotation and negotiation situation. However, any potential political bias in the LLMs might be reflected in our data and model development.

We curated a dataset of labelled political party manifestos. This was done leveraging two existing datasets, the manifesto project~\citep{lehmann2023manifesto} and COALITIONAGREE~\citep{kluver2019coalition}, which both are under CC0 licensing.

\section{Limitation}
The data that we have introduced in this work in limited to six European countries. We believe that same process can be applied to the data from other countries as well. Since we had limited access to experts in the field of political science, we could not perform human annotation on the data that we have collect. We plan to do this in our next steps. In addition, we have considered a scenario where only two political parties start a coalition negotiation, and the process is only limited to discussing the content of manifesto statements. However, in reality this process involves many complex factors, such as seat distribution and political partners, that we could not account for, due to data and model limitation. Moreover, due to budget and time limit, the negotiation process was limited to three rounds of negotiation between the parties.
\bibliography{anthology,custom}

\appendix
\section{Prompts}
\label{sec:prompts}

In this section, we present the prompt template that we have used in our proposed model. The provided tables cover the prompts for calling \texttt{gpt-3.5-turbo}, however, adapted versions of these were used to call \texttt{LLaMa}-based models.
\begin{table*}
    \resizebox{0.7\linewidth}{!}{
    \begin{tabular}{l}
        \hline
        \textbf{Prompt Initial Query} \\
        \hline
        \texttt{system}: Ignore previous instruction. \\
                         You are the representative of the \$PARTY. \\
                         As a representative of the \$PARTY, your job is to negotiate a coalition with \$OPPOSING-PARTY. \\
                         Consider the core values of your party at each step. \\
                         You must think strategically to make this negotiation happen. \\
                         Use diplomatic and political tone in responding. \\
        \texttt{user}: <CONTEXT> \\
\$OPPOSING-PARTY \$ACTION the following statement: \\
statement: \$STATEMENT \\
because \$EXPLANATION
The importance of the statement to your party: \$STATEMENT-SCORE \\
Status of the negotiation so far: \$REFLECTION \\
</CONTEXT> \\
Please think about what your stance is with respect to this statement? \\
Use the following notes and output format \\
<NOTES> \\
- Think strategically. \\
- Choose from SUPPORT or OPPOSE. \\
- Consider how important and how relevant this statement is to your party's core value and platform. \\
- Consider how important and how relevant this statement is to the opposition party's core value and platform. \\
</NOTES> \\
Use the following output format: \\
<REASON>explain what should be your stance about the statement</REASON> \\
<ANSWER>use the reason to decide if you should SUPPORT or OPPOSE the statement</ANSWER> \\
\hline
    \end{tabular}}
\end{table*}

\begin{table*}[]
    \resizebox{0.8\linewidth}{!}{
    \begin{tabular}{l}
    \hline
    \textbf{Prompt Follow Up Query} \\
    \hline
    \texttt{user}: <CONTEXT> \\
       In the previous round of negotiation, while you \$ACTION the following statement, \$OPPOSING-PARTY did not agree with you:  \\
statement: \$STATEMENT because \$OPPOSING-PARTY-EXPLANATION\\
The importance of the statement to your party: \$STATEMENT-SCORE \\
Given the status of the negotiation so far: [SELF-REF] \\
Now you have four options: \\
- support the statement because it is important to your party \\
- oppose the statement because you can make sacrifice to make the agreement happen. 
\\ Choose this if the statement is not that important to your party's core value.\\ 
- refine the statement because you can make sacrifice by proposing a refined version of the statement \\ that can make the final agreement happen. \\
- compromise over another statement because you can see this as a bargaining opportunity. \\
</CONTEXT> \\
Please think about what option you choose this time.\\ 
Use the following notes and output format: \\
<NOTES>\\
- Think strategically. \\
- choose from SUPPORT or OPPOSE or COMPROMISE or REFINE. \\
- Consider how important and how relevant this statement is to your party's core value and platform.\\
- Consider how important and how relevant this statement is to the opposition party's core value and platform. \\
</NOTES>\\

Use the following output format: \\
<REASON>explain what should be your stance about the statement</REASON> \\
<ANSWER>use the reason to decide if you should SUPPORT or OPPOSE or COMPROMISE or REFINE</ANSWER> \\
    \hline
    \end{tabular}}
\end{table*}

\begin{table*}[]
    \resizebox{0.7\linewidth}{!}{
    \begin{tabular}{l}
    \hline
    \textbf{Prompt Ask for Statement Refinement}\\
    \hline
    \texttt{user}: Please think about a refined version of the statement. \\
    Use the following notes and output format: \\
<NOTES> \\
- Think strategically and propose a new version of the statement. \\
- Carefully tailor the statement in a way that the new statement reflects \\
the combined input and priorities of all participating political parties.\\
</NOTES>\\

Use the following output format: \\
<REASON>explain why the refined statement can help the coalition agreement</REASON> \\
<ANSWER>use the reason to generate a new version of the statement.</ANSWER>\\
    \hline

    \end{tabular}}

\end{table*}

\begin{table*}
    \resizebox{0.7\linewidth}{!}{
    \begin{tabular}{l}
    \hline
    \textbf{Prompt Ask for Compromise} \\
    \hline
    \texttt{user}: <CONTEXT> \\
followings are the list of statements from the \$OPPOSING-PARTY that you have have previously did not support: \\
\$LIST \\
</CONTEXT> \\
Please think through the statements, and decide on which one you are willing to change your stance. \\ 
Use the following notes and output format:\\
<NOTES> \\
- Choose a statement that is the least important to your party. \\
</NOTES> \\

Use the following format: \\
<REASON>explain which one of these statements you should change your stance and support</REASON> \\
<ANSWER>use the reason to choose the statement that you support</ANSWER> \\
    \hline

    \end{tabular}}

\end{table*}

\begin{table*}
    \resizebox{0.7\linewidth}{!}{
    \begin{tabular}{l}
    \hline
    \textbf{Prompt Compromise Follow Up} \\
    \hline
    <CONTEXT>
statement: \$COMPROMISE-STATEMENT \\
\$OPPOSING-PARTY party has agreed to support the above statement from your party if you support their statement. \\
Their statement: \$STATEMENT \\
</CONTEXT> \\
Please think about what your stance is with respect to their statement?\\
Use the following notes and output format:\\
<NOTES>\\
- Consider that \$OPPOSING-PARTY changed their stance on another statement to make this agreement happen \\
- Consider your party's core values and platform \\
- Consider how important and how relevant this statement is to your party's core value and platform. \\
- Consider how important and how relevant this statement is to the opposition party's core value and  platform. \\
- Think strategically \\
</NOTES> \\

Use the following output format: \\
<REASON>explain what should be your stance about the statement</REASON> \\
<ANSWER>use the reason to decide if you should SUPPORT or OPPOSE the statement</ANSWER> \\

    \hline

    \end{tabular}}

\end{table*}

\end{document}